%% file: main.tex
\documentclass[10pt,twocolumn,letterpaper]{article}

\usepackage{iccv}              % To produce the CAMERA-READY version
\usepackage[accsupp]{axessibility}  % Improves PDF readability for those with disabilities.

\usepackage{graphicx}
\usepackage{afterpage}
\usepackage{tabularx}
\usepackage{makecell}
\usepackage{amsmath} % For math symbols and commands
\usepackage[ruled]{algorithm2e}
\usepackage{adjustbox}
\usepackage{multirow}
\usepackage{pifont}
\input{preamble}

\definecolor{iccvblue}{rgb}{0.21,0.49,0.74}
\usepackage[pagebackref,breaklinks,colorlinks,allcolors=iccvblue]{hyperref}

\def\paperID{7432} % *** Enter the Paper ID here
\def\confName{ICCV}
\def\confYear{2025}

\title{\algoname: Zero-Shot Unified Image Restoration via Latent Diffusion\\ Recurrent Posterior Sampling}

\author{
    Huaqiu Li\textsuperscript{1,2}\footnotemark[1], Yong Wang\textsuperscript{2}\footnotemark[2] \footnotemark[3], Tongwen Huang\textsuperscript{2}, Hailang Huang\textsuperscript{2}, Haoqian Wang\textsuperscript{1}\footnotemark[2], Xiangxiang Chu\textsuperscript{2}\\
    \textsuperscript{1}Tsinghua University\quad\textsuperscript{2}AMAP, Alibaba Group \\
    \small \texttt{lihq23@mails.tsinghua.edu.cn, wanghaoqian@tsinghua.edu.cn} \\
    \small \texttt{\{wangyong.lz, huangtongwen.htw, huanghailang.hhl, chuxiangxiang.cxx\}@alibaba-inc.com}
    }
\begin{document}
\input{pdf/teasor_fig}
\maketitle
{
\renewcommand{\thefootnote}{\fnsymbol{footnote}}
\footnotetext[1]{\ Work done during the internship at AMAP, Alibaba Group.}
\footnotetext[2]{\ Corresponding author.}
\footnotetext[3]{\ Project lead.}
}
\input{sec/0_abstract}
\input{sec/1_intro}
\input{sec/related_work}
\input{sec/preliminary}

\input{sec/method}
\input{sec/experiment}

%\clearpage
{
    \small
    \bibliographystyle{ieeenat_fullname}
    \bibliography{main}
}
%\clearpage
% \setcounter{page}{1}
% \maketitlesupplementary
% \appendix
%\input{sup.tex}

\end{document}

%% file: preamble.tex
\newcommand{\algoname}{LD-RPS\xspace}
\newcommand{\moduname}{F-PAM\xspace}

%% file: pdf/teasor_fig.tex
% !TEX root = ../PaperForReview.tex

\twocolumn[{%
\renewcommand\twocolumn[1][]{#1}%
\maketitle

\vspace{-1cm}

\begin{center}
    \centering
    % \vspace{-8pt}
    % \setlength{\abovecaptionskip}{0.15cm}
    \captionsetup{type=figure}
    \includegraphics[width=\textwidth]{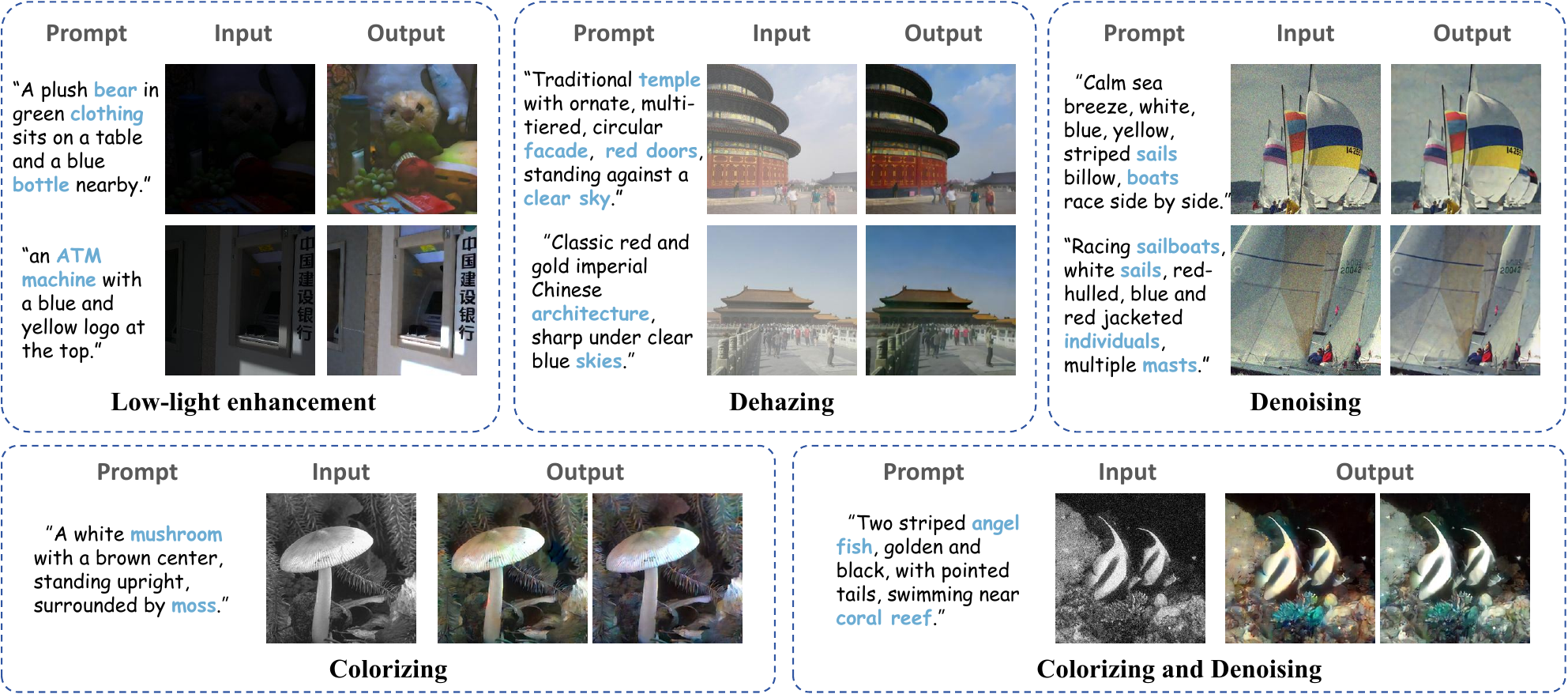}
    \vspace{-2em}
    \captionof{figure}{\algoname has the capability to achieve high-quality \textbf{zero-shot blind restoration} in multiple tasks. Leveraging auxiliary text (keywords highlighted in blue) that describes image content or semantic information, our method achieves superior results in single degradation tasks, including image dehazing, denoising, and colorization, as well as in mixed degradation tasks, including low-light enhancement with denoising and image colorization with denoising.}
    \label{teaser}
    % \vspace{-1em}
\end{center}%
}]

\newcommand\blfootnote[1]{%
\begingroup
\renewcommand\thefootnote{}\footnote{#1}%
\addtocounter{footnote}{-1}%
\endgroup
}

%% file: sec/0_abstract.tex
\begin{abstract}
Unified image restoration is a significantly challenging task in low-level vision. Existing methods either make tailored designs for specific tasks, limiting their generalizability across various types of degradation, or rely on training with paired datasets, thereby suffering from closed-set constraints. To address these issues, we propose a novel, dataset-free, and unified approach through recurrent posterior sampling utilizing a pretrained latent diffusion model. 
Our method incorporates the multimodal understanding model to provide sematic priors for the generative model under a task-blind condition. Furthermore, it utilizes a lightweight module to align the degraded input with the generated preference of the diffusion model, and employs recurrent refinement for posterior sampling. Extensive experiments demonstrate that our method outperforms state-of-the-art methods, validating its effectiveness and robustness. Our code and data will be available at \href{https://github.com/AMAP-ML/LD-RPS}{https://github.com/AMAP-ML/LD-RPS}.
\end{abstract}

%% file: sec/1_intro.tex
\vspace{-4mm}
\section{Introduction}
Image quality is crucial for downstream tasks, like object detection~\cite{object,diwan2023object}, and face recognition~\cite{kim2022adaface,meng2021magface}. During the processes of acquisition, storage, and transformation, images are vulnerable to various degradations, such as noise, low light, and motion blur. Traditional studies~\cite{restormer,prompt-sid,deblurring,SR,RDDM,deraining,dark_channel,chen2025spatiotemporal, transplat} focus on task-specific solutions, employing designed network architectures that are optimized to address particular types of degradation using the corresponding datasets. These methods demonstrate strong performance on their respective tasks while showing limited generalizability to other degradation scenarios.

In recent years, unified image restoration~\cite{Perceiver,transweather,diffuir,bad_weather,prores} has gained significant attention in the research community. This approach aims to develop a single model proficient in handling multiple restoration tasks, thereby improving generalization performance. Traditional methods, such as AirNet~\cite{airnet}, typically utilize datasets with various degradations for training, but lack modules explicitly designed to identify degradation patterns. Subsequent studies, such as ~\cite{promptir,instructir}, incorporate degradation learning modules into network architecture, with this concept further extended to diffusion-based methods~\cite{daclip,autodir}.

However, these data-driven approaches often exhibit limited generalization to degradation types not present in the training datasets~\cite{openset1,openset2,mae}. Furthermore, the creation of comprehensive datasets that encompass diverse types of degradation is both time-intensive and laborious. Therefore, an efficient solution for unified image restoration must optimally satisfy three essential criteria: \textbf{1) employ unsupervised training to minimize dependency on labeled data; 2) be dataset-free to reduce training costs; and 3) generalize effectively to the unseen types of degradation.}

\begin{figure}[tbp]
  \centering
  \includegraphics[width=0.85\linewidth]{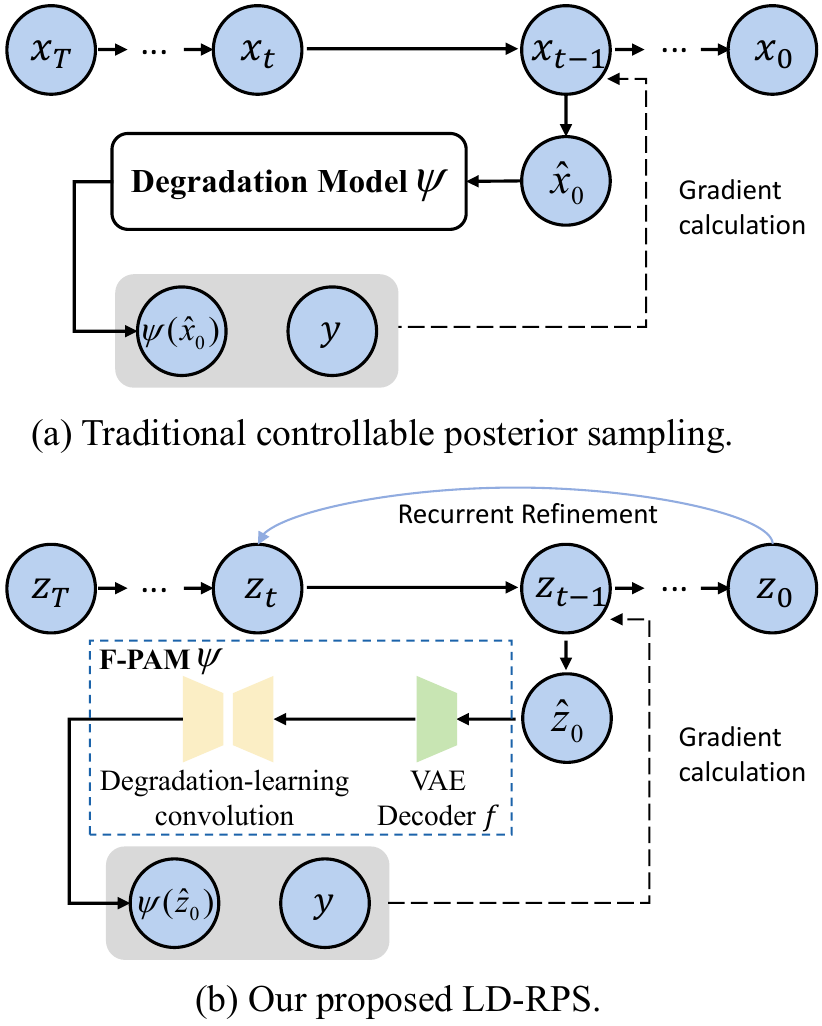}
  \caption{Comparison between traditional diffusion posterior sampling methods for solving inverse problems and our recurrent posterior sampling approach based on latent diffusion.}
  \label{fig:cp}
  \vspace{-1.5em}
\end{figure}

In this paper, we propose \algoname, a unified image restoration method based on latent diffusion and recurrent posterior sampling. This method is dataset-free and operates in an unsupervised, zero-shot manner, utilizing only a single low-quality image during testing. \algoname employs this image as conditional input to guide the diffusion model in generating the corresponding restored image. Leveraging the comprehension capabilities of multimodal large language model (MLLM), \algoname generates textual prompts based on low-quality images, thus providing prior semantic information for image generation. Meanwhile, using feature and pixel alignment module (F-PAM), \algoname aligns intermediate results in reverse diffusion with the degraded image, correcting the direction of posterior sampling via their loss to achieve semantic consistency. We further employ a recurrent strategy to initialize the posterior sampling based on the preliminarily restored image, thereby enhancing the stability of the generative model and refining the results. Extensive experiments demonstrate that \algoname significantly outperforms state-of-the-art methods. 

As shown in Fig.~\ref{fig:cp}, the main differences between our approach and previous posterior sampling methods are: \textbf{the utilization of latent space representation, the adoption of learnable networks to establish domain mapping, and the recurrent bootstrap optimization scheme}.

\textbf{Why use latent diffusion?}
%Variational Autoencoders (VAEs) effectively retain essential features and structural information during the encoding process, inherently filtering out certain unrelated degradation details such as noise and blur. 
Previous representation learning methods have shown that pixel-level image information is often redundant~\cite{vae,mae}, and degraded images have additional meaningless noise. Compressing images into the latent space allows the model to capture essential structures and semantics while filtering out some degradation details like noise and blurriness.
%Moreover, the latent space typically exhibits a smoother distribution of samples~\cite{sd,vae}, leading to the generation of images that are more natural and coherent.

\textbf{Why use a learnable network to model $\psi$?}
In pixel-level diffusion and non-blind methods~\cite{gdp,ddnm}, $\psi$ is often linearly modeled as: $ \boldsymbol{y} = \mathbf{A}\boldsymbol{x} + \mathbf{B}$, where $\boldsymbol{x}$ represents the pixel-level image. 
However, in our case, we need to consider both the alignment of the latent and the image space, as well as the analysis of complex real-world degradations, making explicit modeling unsuitable for this scenario.

So our contributions are summarized as follows:
\begin{itemize}
    \item We propose a multimodal zero-shot unified image restoration framework, \algoname, which leverages the semantic information inherent in degraded images to achieve generative restoration.
    \item To optimize the direction of posterior sampling, we design an unsupervised F-PAM to bridge the gap between degraded images and the generation of latent features.
    \item We develop a recurrent posterior sampling strategy that refines the initialization point of diffusion and progressively enhances image quality.
    \item Our method surpasses state-of-the-art posterior sampling approaches in addressing unified restoration problems.
\end{itemize}

%% file: sec/related_work.tex
\begin{figure*}[tbp]
  \centering
  \includegraphics[width=\linewidth]{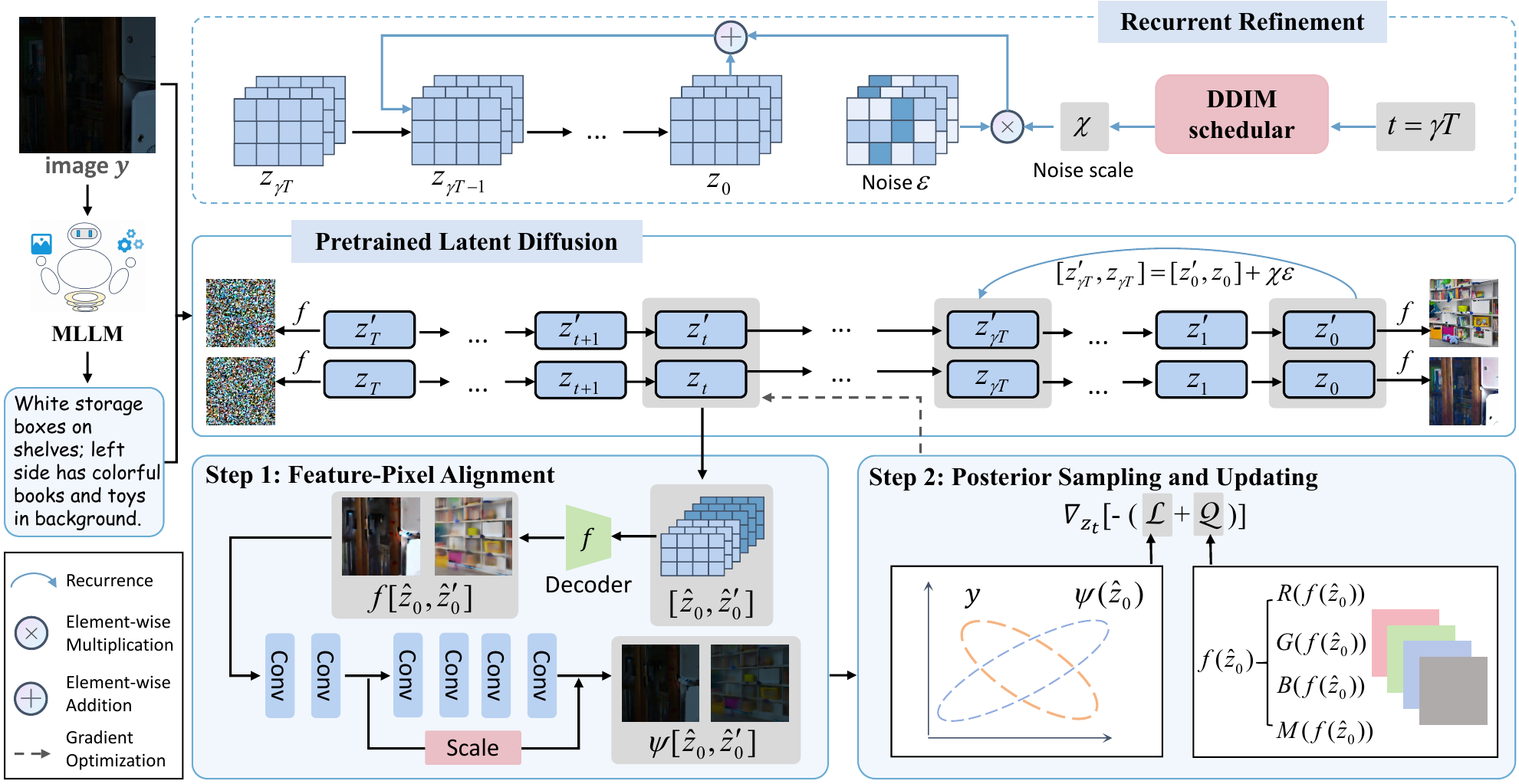}
  \vspace{-1.5em}
  \caption{The overall framework of \algoname. Initially, \algoname utilizes MLLMs to annotate the low-quality image and generate prompts. Based on these prompts, two distinct text-to-image processes are carried out: free diffusion and posterior sampling. In step 1, intermediate data produced by the diffusion process are employed to train and infer \moduname, aligning the diffusion feature domain with the degraded image domain. In step 2, distance loss and quality loss are computed using the output of \moduname and the intermediate diffusion results, with gradients propagated back. The entire diffusion process is recurrently conducted in a bootstrap manner to enhance generation quality. In the figure, $R$, $G$, $B$, and $M$ represent the three image channels and their mean, respectively.}
  \label{fig:pipe}
  \vspace{-1em}
\end{figure*}

\section{Related Work}
\noindent \textbf{Unified image restoration.} Unified image restoration (UIR) aims to address multi-type restoration tasks by developing a single robust model. These methods are categorized into blind and non-blind approaches based on the incorporation of prior knowledge regarding degradation types. Non-blind methods~\cite{nb1, nb2, instructir,diffbir,ddnm} utilize these priors to explicitly address specified degradations. In contrast, blind restoration techniques~\cite{b1,b2,autodir,diffuir,airnet,daclip,huaqiu_interpretable} handle unknown degradations without relying on explicit prior knowledge of the degradation types. This presents considerable challenges for practical applications, which often involve non-linear and composite degradation patterns.

In the realm of UIR, when the trainset and testset are identical and comprise a single image, the approach is identified as zero-shot. Previous works~\cite{diffuir,Perceiver} discuss the zero-shot generalization capability but are limited to certain datasets. Furthermore, studies including \cite{ddnm,TAO,diffbir,gdp} leverage the knowledge from pre-trained diffusion models to facilitate zero-shot generalization. However, these approaches often rely on assumpted prior or produce restoration results with considerable randomness and instability.

\noindent \textbf{Diffusion application.} Diffusion models~\cite{ddpm,sde,sdedit,palette,sd,dalle,fluxtext, usp} have significantly advanced low-level vision tasks and opened new avenues for innovative approaches to image restoration.  DA-CLIP~\cite{daclip} achieves blind degradation recognition by initially fine-tuning the CLIP model, followed by fine-tuning a pre-trained diffusion model to establish the unified restoration. The technique of test-time adaptation~\cite{tta2,tta3,mmtta} is frequently employed for the reverse process of diffusion, facilitating the progressive adaptation of intermediate states to conditional information during inference. However, these approaches face the challenge of balancing the generative capability of diffusion with the semantic coherence introduced by conditional information~\cite{diffusion_beat_gans,controlnet}, which inevitably affects the similarity between the restored image and the ground truth.

%% file: sec/preliminary.tex
%\vspace{-1em}
\section{Preliminary}
\noindent \textbf{Inference of the latent diffusion.} The proposed \algoname method leverages latent diffusion~\cite{sd} to perform the inference process, which involves the iterative generation of distinct image features from the initial noise, followed by the decoding of these features into standard images. Suppose that the latent representation of the image, denoted as $\mathbf{z}$, follows the distribution $\mathbf{z} \sim q_1(\mathbf{z})$. Within text-to-image generation, the timesteps are discretized into $T$ intervals. At the time step $T$, we sample $\mathbf{z}_T \sim \mathcal{N}(\mathbf{0}, \mathbf{I})$, which serves as the starting point for the reverse diffusion process. Let $\epsilon_{\theta}$ represent the output of the noise prediction network, and let $\mathbf{c}$ denote the text embedding. This process can be formally expressed using the following equations:
%\vspace{-3mm}
\begin{equation}
\mathbf{z}_{t-1} \sim q(\mathbf{z}_{t-1} | \mathbf{z}_t) = \mathcal{N}(\mathbf{z}_{t-1};\mu(\mathbf{z}_t, \hat{\mathbf{z}}_0), \sigma_t^2\mathbf{I}),
\end{equation}
%\vspace{-4mm}
\begin{equation}
    \mu(\mathbf{z}_t, \hat{\mathbf{z}}_0 ) = \frac{\sqrt{\Bar{\alpha}_{t-1}}\beta_t \hat{\mathbf{z}}_0 + \sqrt{\alpha_t}(1-\Bar{\alpha}_{t-1}) \mathbf{z}_t}{1 - \Bar{\alpha}_t},
\end{equation}
\begin{equation}\small
   \hat{\mathbf{z}}_0 = \frac{\mathbf{z}_t - \sqrt{1 - \Bar{\alpha}_t} \epsilon_{\theta}(\mathbf{z}_t,t,\mathbf{c})}{\sqrt{\Bar{\alpha}_t}}, 
\; \sigma_t^2 = \frac{1-\Bar{\alpha}_{t-1}}{1-\Bar{\alpha}_t}\beta_t, 
\label{equ:x0}
\end{equation}
where $t \in \{0, ..., T\}$, $\Bar{\alpha}_t=\prod^t_{i=0}\alpha_i$, $\alpha_i = 1-\beta_i$ and $\beta_i$ denotes the variance at the $i$-th timestep. The denoised feature $\mathbf{z}_0$ is subsequently passed through the VAE decoder $f$ to produce the final RGB image $\boldsymbol{x}_0 = f(\mathbf{z}_0)$.

\noindent \textbf{Controllable posterior sampling.} %For the image restoration, the relationship between a low-quality image $\boldsymbol{y}$ and its corresponding ground truth $\boldsymbol{x}$ can be formulated as $\boldsymbol{y}=\mathcal{R}(\boldsymbol{x})$. 
When performing a zero-shot restoration for $ \boldsymbol{y} $, which belongs to a distribution $\boldsymbol{y}\sim q_2(\boldsymbol{y})$ we incorporate it as a conditional input to ensure semantic consistency during the reverse diffusion process. Therefore, the problem of posterior estimation $p_\theta(\mathbf{z}_{t-1}|\mathbf{z}_{t})$ can be transformed into $ p_\theta(\mathbf{z}_{t-1}|\mathbf{z}_{t}, \boldsymbol{y}) $~\cite{Deep_unsupervisedlearning,diffusion_beat_gans}:
\vspace{-1mm}
\begin{equation}
\begin{split}
    \log p_\theta\left(\mathbf{z}_{t-1}|\mathbf{z}_{t},\boldsymbol{y}\right) &\!=\!\log \left(p_\theta\left(\mathbf{z}_{t-1}|\mathbf{z}_{t}\right) p\left(\boldsymbol{y}| \mathbf{z}_{t}\right)\right)\!+\!C_1 
    \\ \!\!\!\!  & \approx \log q(\boldsymbol{r})+C_2, 
\end{split}
\end{equation}
%\vspace{-1mm}
where $\boldsymbol{r} \sim \mathcal{N}(\boldsymbol{r} ; \mu(\mathbf{z}_t, \hat{\mathbf{z}}_0)+\delta \boldsymbol{g}, \delta)$. Here, $\boldsymbol{g}=\nabla_{\mathbf{z}_{t}} \log p(\boldsymbol{y} \mid \mathbf{z}_{t})$, and variance $\delta = \delta_{\theta}\left(\mathbf{z}_{t}\right)$, which can be fixed
to a known constant. The parameters $C_1$ and $C_2$ are constants. The term $p\left(\boldsymbol{y} \mid \mathbf{z}_t\right)$ denotes the probability of obtaining the corresponding GT to the degraded image $\boldsymbol{y}$ by denoising $\mathbf{z}_t$, which can be estimated by $p\left(\boldsymbol{y} \mid \hat{\mathbf{z}}_0\right)$. This term can be further derived as:
\vspace{-1mm}
\begin{equation}
\begin{split}
    p\left(\boldsymbol{y} \mid \hat{\mathbf{z}}_0\right) = \frac{1}{Z} & \exp \left(-\left[\mathcal{L}\left(\psi(\hat{\mathbf{z}}_0), \boldsymbol{y}\right) + \mathcal{Q}(\hat{\mathbf{z}}_0) \right]\right),
\end{split}
\label{equ:p_y_x0}
\end{equation}
%\vspace{-1mm}
where $\psi$ represents the function that maps $q_1(\mathbf{z})$ to $q_2(\boldsymbol{y})$, $\mathcal{L}$ represents a certain image distance metric, such as MSE, and $\mathcal{Q}$ is responsible for evaluating the quality of the decoded image corresponding to $\hat{\mathbf{z}}_0$. The term $Z$ is a scale factor. Through the above derivation, we convert the problem of conditional posterior sampling $p_\theta\left(\mathbf{z}_{t-1}|\mathbf{z}_{t},\boldsymbol{y}\right)$ into the task of solving $\nabla_{\mathbf{z}_{t}} \log p(\boldsymbol{y} \mid \hat{\mathbf{z}}_0)$, which involves computing the gradient of the total composite loss $-\left[\mathcal{L}\left(\psi(\hat{\mathbf{z}}_0), \boldsymbol{y}\right) + \mathcal{Q}(\hat{\mathbf{z}}_0) \right]$ with respect to $\mathbf{z}_t$.

%% file: sec/method.tex
\section{Method}
Our proposed \algoname introduces a novel framework for restoring a degraded image without requiring training or any prior input. We leverage the comprehension capabilities of MLLMs to extract semantic priors from the original low-quality images and use them to guide the recurrent posterior sampling process. To achieve controllable posterior sampling using a learnable network to model $\psi$, as shown in Fig.~\ref{fig:pipe}, we perform \textbf{two steps} for each iteration of the reverse process in latent diffusion. Moreover, we integrate the bootstrap concept in \algoname by incorporating \textbf{a recurrent mechanism} into the unidirectional posterior sampling process, repeatedly utilizing the high-quality images already restored to improve the generative results.

%\noindent\textbf{Step 1}: unsupervised optimization of the DL-Adapter. 

%\noindent\textbf{Step 2}: using the adapter aligning the diffusion model's generated domain with the degraded domain, and updating the intermediate states $\mathbf{z}_t$ with the respective losses. 
%Our proposed \algoname introduces a novel framework for restoring a degraded image without requiring the training on a dataset. To fully harness the capabilities of diffusion modelically generated by processing the degraded input via a MLLM model, which analyzes visual features and formulates semantically relevant textual cues. The technical implementation of this prompt generation pipeline is elaborated in Section \ref{PGP}.

%As previously discussed, the critical challenge in controllable posterior sampling lies in solving Eq.~\ref{equ:p_y_x0}, where $\mathcal{R}$ represents the process of image degradation. To address this issue, we propose DL-Adapter to implicitly learn the degradation patterns rather than relying on explicit modeling. The DL-Adapter is simultaneously optimized with the posterior sampling process of latent diffusion.
%However, the degradation processes in real world is commonly intricate and non-linear, deviating from the distribution illustrated by the above equation. Therefore, employing a learning-based approach to represent degradation facilitates the capture of the complex mappings inherent in such processes.

\subsection{Task-Blind Semantic Prior Generation}
\label{PGP}
To ensure robust generation for latent diffusion posterior sampling, we utilize text embeddings $\boldsymbol{c}$ for classifier guidance, effectively steering the model towards generating the target content. Ideally, the textual prompt should convey accurate image information based on human preferences, free from any degradation-related noise.

 We aim to acquire sufficient semantic priors to initialize the posterior sampling process of the generative model. Due to the rapid development of large-language-model~\cite{huang2025adaptive,huang2026does,huang2026real,zhang2026heterogeneous}, we compensate for the lack of prior information in the model input through the comprehension capabilities of MLLMs. Consequently, by merely inputting low-quality images and manually crafted prompts, we can obtain the desired content. The crafted prompt is structured as follows:
\begin{figure}[htbp]
  \centering
  %\vspace{-2mm}
\includegraphics[width=\linewidth]{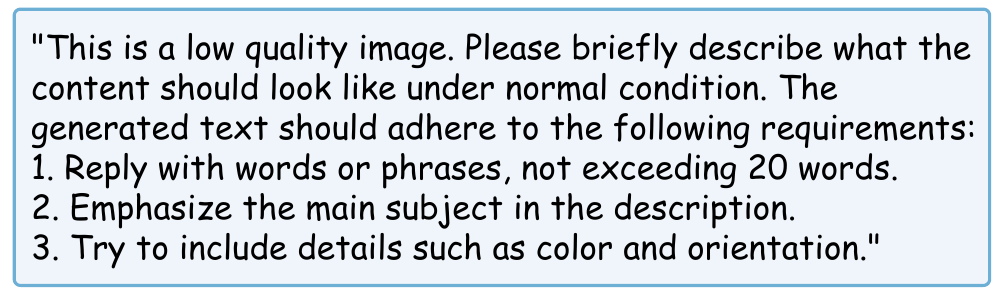}
\end{figure}

% \ywang{The term in \textbf{bolded words} can be replaced with alternative task classifications in the context of handling low-quality images. It is crucial to emphasize that the descriptions related to degradation types are employed for generating high-quality text and are not utilized as priors for optimization. Consequently, our method remains as a blind restoration approach.}
\vspace{-3mm}
\subsection{Feature and Pixel Alignment Module}
%\subsection{Degradation-Learning Adapter}
%结构、输入输出、优化损失、优化阶段
\textbf{Motivation.} To prevent identity mapping, we conduct two concurrent reverse diffusion processes: one utilizes an intermediate result modified by $\nabla_{\mathbf{z}_{t}} \log p(\boldsymbol{y} \mid \hat{\mathbf{z}}_0)$ to yield a new $\mathbf{z}_t$, while the other only guided by text embeddings to generate an intermediate $\mathbf{z}_t'$. We concatenate them as $[\mathbf{z}_t,\mathbf{z}_t']$, and then derive $\left[\hat{\mathbf{z}}_0,\hat{\mathbf{z}}_0'\right]$ by Equ.~\ref{equ:x0}. We need to align $\mathbf{z}_t$ with the conditional information $\boldsymbol{y}$ at each iteration to achieve controllable posterior sampling. The distinctions between these two distributions involve \textit{the space gap, which is the difference between the latent space and the image space}, and \textit{the domain gap, which is the difference between the normal image domain and the degraded domain}. Adopting an unsupervised strategy to fit both requires accurately aligning their gaps and designing corresponding losses. Therefore, we propose the F-PAM network, which consists of a frozen VAE~\cite{vae} decoder $f$ and degradation-learning convolutions.
%\vspace*{-5mm}
\input{tab/notation}
%We employ the DL-Adapter to implicitly model degradation $\mathcal{R}$, thereby mitigating the domain gap between the degraded image $\boldsymbol{y}$ and generated normal image $\hat{\boldsymbol{x}}_0$. To achieve this, we define the reverse diffusion process with $T$ timesteps, and $t \in \{0, 1, 2, \ldots, T\}$. To prevent identity mapping, we conduct two concurrent reverse diffusions: one utilizes an intermediate result modified by $\nabla_{\mathbf{z}_{t}} \log p(\boldsymbol{y} \mid \hat{\boldsymbol{x}}_0)$ to yield a new $\mathbf{z}_t$, while the other employs diffusion only guided by text embeddings to generate an intermediate $\mathbf{z}_t'$. Following Eq.~\ref{equ:x0}, we calculate $\hat{\mathbf{z}}_0$ and $\hat{\mathbf{z}}_0'$. These terms are then decoded using the VAE as $\hat{\mathbf{x}}_0= f_{\theta}(\hat{\mathbf{z}}_0)$ and $\hat{\mathbf{x}}_0'=f_{\theta}(\hat{\mathbf{z}}_0')$, respectively. The concatenation of these decoded representations, denoted as $ [\hat{\mathbf{x}}_0,\hat{\mathbf{x}}_0'] $, is subsequently processed by the DL-Adapter.

This network simulates a degradation process applied to the concatenated image:
\begin{equation}     \psi[\hat{\mathbf{z}}_0,\hat{\mathbf{z}}_0'] = h_2(h_1(f[\hat{\mathbf{z}}_0,\hat{\mathbf{z}}_0'])) + \mathbf{p}\odot h_1(f[\hat{\mathbf{z}}_0,\hat{\mathbf{z}}_0']),
\end{equation}
where $ \psi $ represents the F-PAM, $f$ represents the decoder, $ h_1 $ and $ h_2 $ represent the convolutional networks, and $\mathbf{p}$ denotes a learnable channel attention factor. The \moduname is optimized concurrently with the reverse diffusion process by using the following loss function~\cite{TAO}:
\vspace{-2mm}
\begin{equation}
    \begin{aligned}
    \mathcal{S}_{\psi}&=\lambda_1\parallel f[\hat{\mathbf{z}}_0,\hat{\mathbf{z}}_0'] - \psi[\hat{\mathbf{z}}_0,\hat{\mathbf{z}}_0'] \parallel_2^2 \\
    &+\lambda_2\parallel \mathcal{V}(f[\hat{\mathbf{z}}_0,\hat{\mathbf{z}}_0']) - \mathcal{V}(\psi[\hat{\mathbf{z}}_0,\hat{\mathbf{z}}_0']) \parallel_2^2 \\
    &+\lambda_3\log(1-\mathcal{D}_1(\psi[\hat{\mathbf{z}}_0,\hat{\mathbf{z}}_0']),
\end{aligned}
\end{equation}
\vspace{-1mm}
where $\mathcal{D}_1$ is a discriminator optimized through~\cite{gan}:
\begin{equation}
   \mathcal{S}_{dis} = -\log \mathcal{D}_1[y, y] - \log(1-\mathcal{D}_1(\psi[\hat{\mathbf{z}}_0,\hat{\mathbf{z}}_0'])),
\end{equation}
$\mathcal{V}(\cdot)$ represents the feature maps extracted from the pretrained perceptional network~\cite{vgg}, while $\lambda_1$, $\lambda_2$, and $\lambda_3$ denote the respective loss weights.

\subsection{Posterior Sampling and Updating}
In the posterior sampling process, the distribution $q(\mathbf{z}_{t-1} | \mathbf{z}_t, \boldsymbol{y}) \propto \mathcal{N}(\mathbf{z}_{t-1}; \mu(\mathbf{z}_t, \hat{\mathbf{z}}_0) + \delta \boldsymbol{g}, \delta),$ is optimized iteratively through a two-stage procedure. Initially, from $ T $ to $ t_1 $, the term $\boldsymbol{g}$ is set to zero, simplifying the distribution as $ q(\mathbf{z}_{t-1} | \mathbf{z}_t, \boldsymbol{y}) = q(\mathbf{z}_{t-1} | \mathbf{z}_t) $. This phase focuses exclusively on optimizing the \moduname to guarantee its rapid and accurate convergence during the early sampling stages. Subsequently, from $ t_1 $ to $ 0 $, a joint optimization of both the \moduname and the posterior estimation is conducted using the formulation $q(\mathbf{z}_{t-1} | \mathbf{z}_t, \boldsymbol{y}) \propto \mathcal{N}(\mathbf{z}_{t-1}; \mu(\mathbf{z}_t, \hat{\mathbf{z}}_0) + \delta \boldsymbol{g}, \delta) $. Within this process, we define the following $\mathcal{L}$ as~\cite{gdp,TAO}: 
\vspace{-1mm}
\begin{equation}
\begin{aligned}
    \mathcal{L}(\psi(\hat{\mathbf{z}}_0),\boldsymbol{y})&=w_1\parallel \boldsymbol{y} - \psi(\hat{\mathbf{z}}_0) \parallel_2^2 \\
    &+w_2\parallel \mathcal{V}(\boldsymbol{y}) - \mathcal{V}(\psi(\hat{\mathbf{z}}_0)) \parallel_2^2 \\
    &+w_3log(1-\mathcal{D}_2(f(\hat{\mathbf{z}}_0) - \boldsymbol{y})),
\end{aligned}
\end{equation}
%\vspace{-3mm}
where $\mathcal{V}$ represents the feature map obtained from the perception network, and $ \mathcal{D}_2 $ represents the type discriminator~\cite{gan}, which iterates with the posterior sampling process:
\vspace{-1mm}
\begin{equation}
\begin{aligned}
    \mathcal{L}_{dis} = - &\log(1-\mathcal{D}_2(f(\hat{\mathbf{z}}_0)-\boldsymbol{y}))\\
        - &\log\mathcal{D}_2(f[\hat{\mathbf{z}}_0,\hat{\mathbf{z}}_0']-\psi[\hat{\mathbf{z}}_0,\hat{\mathbf{z}}_0']).%引一下gan
\end{aligned}
\end{equation}
Simultaneously, we propose to quantify the image quality degradation by evaluating luminance and chrominance:
\vspace{-2mm}
\begin{equation}
    \mathcal{Q}(\hat{\mathbf{z}}_0)\!=\! w_{4}\frac{1}{K}\sum_{i=1}^{K}\left | f(\hat{\mathbf{z}}_0)\!-\!e \right |\!+\!w_{5}\!\!\!\sum_{\forall(p,q)\in \Omega}^{}\!\!\!(V_p\!-\!V_q)^2
\end{equation}
The loss $\mathcal{Q}(\hat{\mathbf{z}}_0)$ enforces constraints on the average brightness of image patches and the overall chromaticity of the image, where $\Omega =\{(R, G),(R, B),(G, B)\}$, $V_p$ denotes the average intensity value of $p$ channel in the generated image, and $e$ represents the exposure standard that aligns with natural perception.

%\vspace{-2em}
\subsection{Recurrent Refinement}
Following posterior sampling for a single image, we obtained preliminary restoration outcomes. However, some results exhibit color casts, artifacts, and other defects. These problems stem from the divergence between the pre-trained model and the target distribution in zero-shot generation. We observe that diffusion models tend to produce results that are more in line with the pre-trained dataset (e.g., animals), as illustrated in the supplementary materials.

To eliminate these effects, inspired by \textbf{the bootstrapping method} in classical machine learning, we design a framework where the results generated in the previous recurrence are used as the initialization for the next recurrence, continuously strengthening through iterative cycles. Specifically, given the total number of recurrences $n$ and the refinement factor $\gamma \in (0, 1)$, we generate the noisy image through a forward diffusion process of the restored image $\boldsymbol{x}_0^{(i)}$ from the $i$-th recurrence to $\gamma T$ timesteps and subsequently transform it into a feature space. Based on these noisy features, we conduct a recurrent posterior estimation. The comprehensive procedure is outlined in Algorithm~\ref{algo}.
%\vspace{-4mm}
\input{sec/algo}

%% file: tab/notation.tex
%\vspace{-3mm}
\begin{table}[htbp]
\begin{adjustbox}{max width=\linewidth}
\begin{tabular}{ccc}
\noalign{\hrule height 1pt}
Notation{\rule{0pt}{1.2em}} & Description & Shape \\ \noalign{\hrule height 1pt}
$[\hat{\mathbf{z}}_0,\hat{\mathbf{z}}_0']$, $\hat{\mathbf{z}}_0$        &features in latent space{\rule{0pt}{1em}}             &$(*,4,\frac{H}{4},\frac{W}{4})$      
\\
$f[\hat{\mathbf{z}}_0,\hat{\mathbf{z}}_0']$, $f(\hat{\mathbf{z}}_0)$        & images in normal domain           &$(*,3,H,W)$       \\
$\psi[\hat{\mathbf{z}}_0,\hat{\mathbf{z}}_0']$, $\psi(\hat{\mathbf{z}}_0)$        &images in degraded domain             &$(*,3,H,W)$       \\ \noalign{\hrule height 1pt}
\end{tabular}
\end{adjustbox}
%\vspace{-2mm}
\caption{Important notations. The symbol ``$*$'' is batch size, representing $1$ for $\hat{\mathbf{z}}_0$ and $2$ for $[\hat{\mathbf{z}}_0,\hat{\mathbf{z}}_0']$, respectively.}
\end{table}
%\vspace{-3mm}

%% file: sec/algo.tex
\vspace{-0.5em}
\begin{algorithm}[h]
\small
	\caption{Recurrent Refinement}
	% \caption{Input the number of iterations $n$, low-quality image $y$, pre-trained diffusion model, and output the restored image $\boldsymbol{x}_0^{(n)}$.}
	\KwIn{Recurrence $n$, degraded image $\boldsymbol{y}$, gradient scale $\delta$, \moduname $\mathcal{\psi}$, distance loss $\mathcal{L}$, image quality loss $\mathcal{Q}$, VAE encoder $\phi$, VAE decoder $f$}
	\KwOut{Output image $\boldsymbol{x}_{0}^{(n)}$ conditioned on $\boldsymbol{y}$}
    \For{$\boldsymbol{i}=0$ \KwTo $n$}{
    \uIf{$\boldsymbol{i}=0$}{
    $\gamma=1,\mathbf{z}_{\gamma T}^{(i)}=\mathbf{z}_{ T}^{(0)}\sim \mathcal{N}(0,\mathbf{I})$
    }
    \uElse{
    $\mathbf{z}_0^{(i-1)}\!\!=\!\phi(\mathbf{\boldsymbol{x}}_0^{(i-1)}),\mathbf{z}_{\gamma T}^{(i)} \!=\!\! \sqrt{\Bar{\alpha}_t} \mathbf{z}_0^{(i-1)}\! +\! \sqrt{1 \!-\! \Bar{\alpha}_t} \epsilon$
    }
    \For{$t=\gamma T$ \KwTo $1$}{
    %$\mu, \Sigma = \mu_{\theta}\left(\mathbf{z}_{t}^{(i)} \right), \Sigma_{\theta}\left(\mathbf{z}_{t}^{(i)}\right)$\\
    $\hat{\mathbf{z}}_0^{(i)} = \frac{\mathbf{z}_t^{(i)} - \sqrt{1 - \Bar{\alpha}_t} \epsilon_{\theta}(\mathbf{z}_t^{(i)},t,\mathbf{c})}{\sqrt{\Bar{\alpha}_t}}$\\
    $\mu(\mathbf{z}_t^{(i)}, \hat{\mathbf{z}}_0^{(i)} ) = \frac{\sqrt{\Bar{\alpha}_{t-1}}\beta_t \hat{\mathbf{z}}_0^{(i)} + \sqrt{\alpha_t}(1-\Bar{\alpha}_{t-1}) \mathbf{z}_t^{(i)}}{1 - \Bar{\alpha}_t}$\\
    $\mathcal{L}_{total}=-[\mathcal{L}(\psi(\hat{\mathbf{z}}_0^{(i)}),\boldsymbol{y}) + \mathcal{Q}(\hat{\mathbf{z}}_0^{(i)})]$\\
    Sample $\mathbf{z}_{t-1}^{(i)}$ by $\mathcal{N}(\mu(\mathbf{z}_t^{(i)}, \hat{\mathbf{z}}_0^{(i)})\!+\!\delta\nabla_{\mathbf{z}_t}\mathcal{L}_{total},\delta)$
    }
    $\boldsymbol{x}_0^{(i)}=f(\mathbf{z}_0^{(i)})$
    }
    \Return $\boldsymbol{x}_0^{(n)}$
\label{algo}

\end{algorithm}
\vspace{-2em}

%% file: sec/experiment.tex
\input{tab/lolv1}
\begin{figure*}[htbp]
  \centering
  \includegraphics[width=\linewidth]{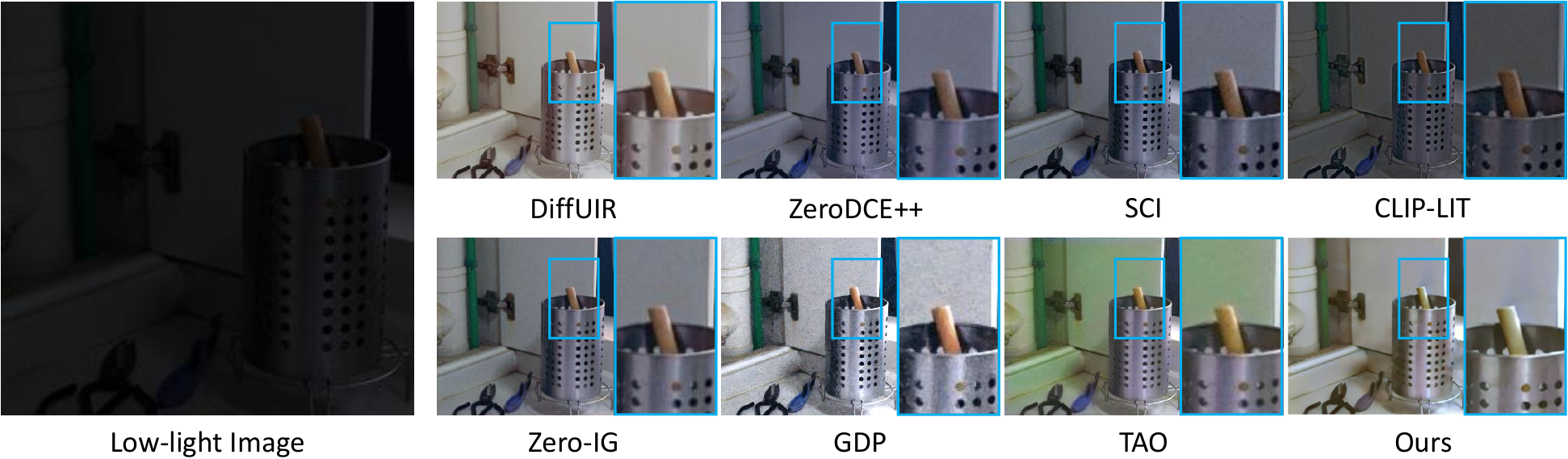}
  \vspace{-5mm}
  \caption{Qualitative comparison results on the LOL dataset are visualized, with details highlighted in blue boxes for closer observation.}
  \label{fig:lol}
  \vspace{-3mm}
\end{figure*}

%\vspace{-2em}
\section{Experiments}
%\subsection{Implementation Details}
%In all experiments, \algoname consistently uses the pre-trained stable diffusion model~\cite{sd}, with sampling conducted via the DDIM scheduler~\cite{ddim}. The time step $T$ is set to 1000, which is divided into 450 sampling steps. The interval where $T > 700$ constitutes the first stage, during which the adapter is trained independently. The second stage is defined by a threshold of $T = 150$, where the quality function is introduced for $T < 150$. Each experiment is conducted using a single Nvidia H20 GPU. Considering the potential impact of initial sampling noise on text-to-image models, which introduces a certain degree of randomness, we average the results obtained from three different random seeds to ensure consistency in the experiments. To enable comparison across various methods, a $256 \times256$ patch is cropped from the center of each image in the test set.

%For the selection of comparison methods, we classify them into three categories: unified supervised methods, task-specific unsupervised methods, and zero-shot methods using posterior sampling. In the comparison of unified supervised methods, we employ AirNet~\cite{airnet}, PromptIR~\cite{promptir}, and DiffUIR~\cite{diffuir}. For the evaluation of zero-shot methods, we utilize GDP~\cite{gdp} and TAO~\cite{TAO}.

\begin{figure*}[htbp]
  \centering
  \includegraphics[width=0.95\linewidth]{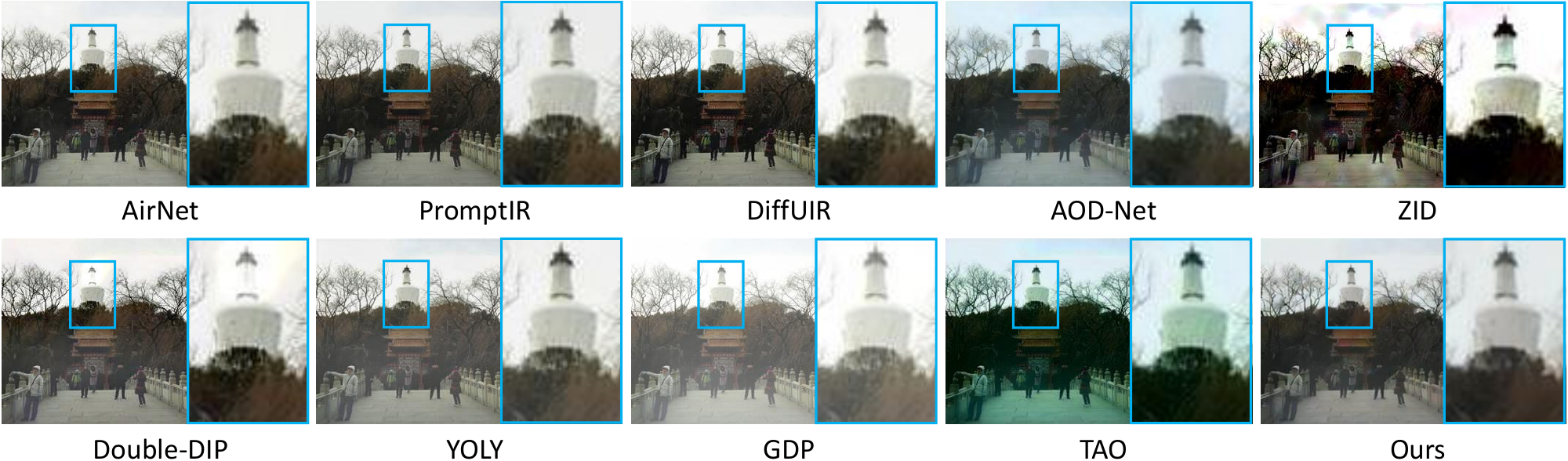}
  \vspace{-0.7em}
  \caption{Qualitative comparison results on the HSTS subset of the RESIDE dataset are visualized.}
  \label{fig:hsts}
  \vspace{-1em}
\end{figure*}
\begin{figure*}[htbp]
  \centering
  \includegraphics[width=0.95\linewidth]{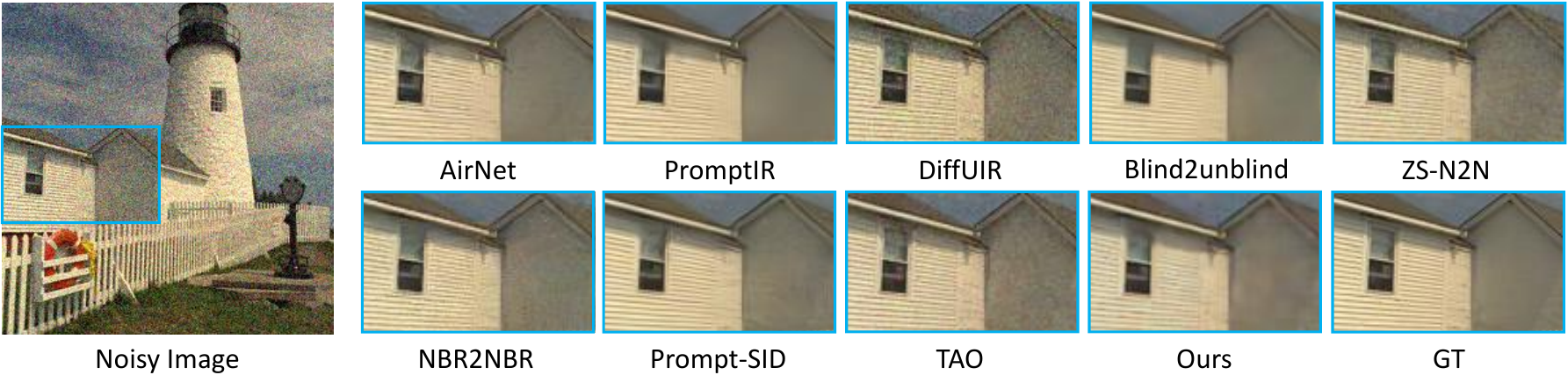}
  \vspace{-0.7em}
  \caption{Qualitative comparison results on the Kodak24 dataset are visualized, with details highlighted in blue boxes for closer observation.}
  \label{fig:kodak}
  \vspace{-1em}
\end{figure*}

\subsection{Experimental Results}
To ensure fairness, all experiments are conducted on a NVIDIA H20 GPU. Due to the stochastic nature of diffusion models, we average the results from three randomly selected seeds to mitigate sensitivity to seed variations. %We use five metrics to evaluate restoration quality: PSNR, SSIM~\cite{ssim}, LPIPS~\cite{lpips}, PI~\cite{pi}, and NIQE~\cite{niqe}. 
More details can be found in the supplementary materials.%加引用We use PSNR, SSIM, and LPIPS metrics to measure the discrepancy between the generated images and the ground truth. For the enhancement task, we also use PI and NIQE to assess image quality. More details can be found in the supplementary materials.

\noindent \textbf{Enhancement.} For this task, we employ ZDCE++~\cite{zdce}, SCI~\cite{sci}, CLIP-LIT~\cite{cliplit}, and Zero-IG~\cite{zeroig} as our task-specific comparison methods. The first three approaches are unsupervised and based on dataset training, whereas Zero-IG is a zero-shot method. As shown in Tab.~\ref{tab:lol} and Fig.~\ref{fig:lol}, our \algoname achieves state-of-the-art results in zero-shot posterior sampling on both LOLv1 and LOLv2~\cite{lolv1} datasets. Furthermore, \algoname surpasses all task-specific methods in PSNR and SSIM metrics and demonstrates comparable performance across other evaluated metrics.

Additionally, we note that although supervised methods DiffUIR~\cite{diffuir} exhibit a greater advantage on full-reference metrics (with ground truth contributing to higher PSNR and SSIM scores), it does not show significant improvements on no-reference metrics such as PI and NIQE when compared to our method. Moreover, due to the closed-set nature of these supervised methods, the models fail to generalize to other data that are not present during training. This limitation also renders AirNet~\cite{airnet} and PromptIR~\cite{promptir} unsuitable for low-light enhancement tasks.

\input{tab/hsts}
\input{tab/denoising}
\input{tab/iter}
\input{tab/text}
\begin{figure}[htbp]
  \centering
  \includegraphics[width=\linewidth]{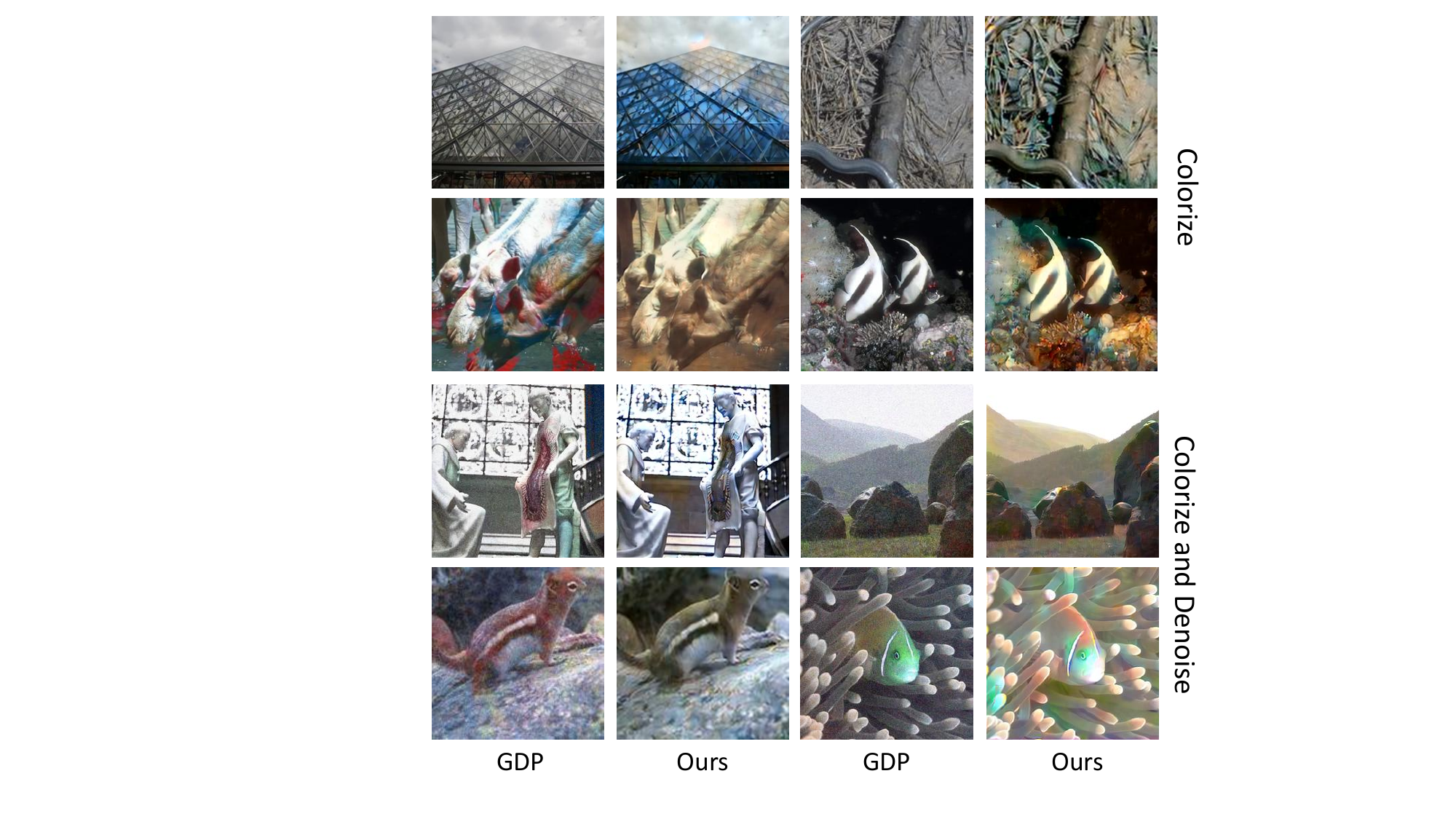}
  \vspace{-5mm}
  \caption{Visualization comparison of \algoname and GDP methods on the image colorization task, as well as the combined task of image colorization and denoising.}
  \label{fig:color}
  \vspace{-1.5em}
\end{figure}

\noindent \textbf{Dehazing.} 
We select AOD-Net~\cite{aod-net}, ZID~\cite{zid}, DDIP~\cite{ddip}, and YOLY~\cite{yoly} as task-specific comparison methods. Among these, AOD-Net is a supervised method utilizing dataset training, whereas the other three methods use zero-shot techniques. Our evaluations were conducted on the HSTS subset of the RESIDE dataset~\cite{reside}. The original GDP framework lacks explicit modeling for degradation patterns in image dehazing. Therefore, we reproduced its dehazing capability by incorporating the modeling of low-light enhancement, which leverages the similarity of global degradation characteristics between the two tasks. As shown in Tab.~\ref{tab:hsts} and Fig.~\ref{fig:hsts}, our \algoname outperforms all zero-shot methods in terms of the PSNR metric.

\noindent \textbf{Denoising.}
To ensure a comprehensive and robust evaluation, we employ the task-specific comparison methods Blind2Unblind~\cite{b2un}, NBR2NBR~\cite{neighbor}, Prompt-SID~\cite{prompt-sid}, and ZS-N2N~\cite{zs-n2n}. Detailed results are presented in Tab.~\ref{tab:kodak} and Fig.~\ref{fig:kodak}. It is important to highlight that GDP fails to precisely model the denoising task, leading to its inability to effectively enhance noisy images. Similarly, DiffUIR~\cite{diffuir} exhibits suboptimal performance due to the lack of noisy-clean paired images in its training set. In contrast, our \algoname consistently outperforms the baseline TAO~\cite{TAO} method across all evaluated metrics.
%\vspace{-2mm}

\begin{figure}[t!]
  \centering
  \includegraphics[width=0.95\linewidth]{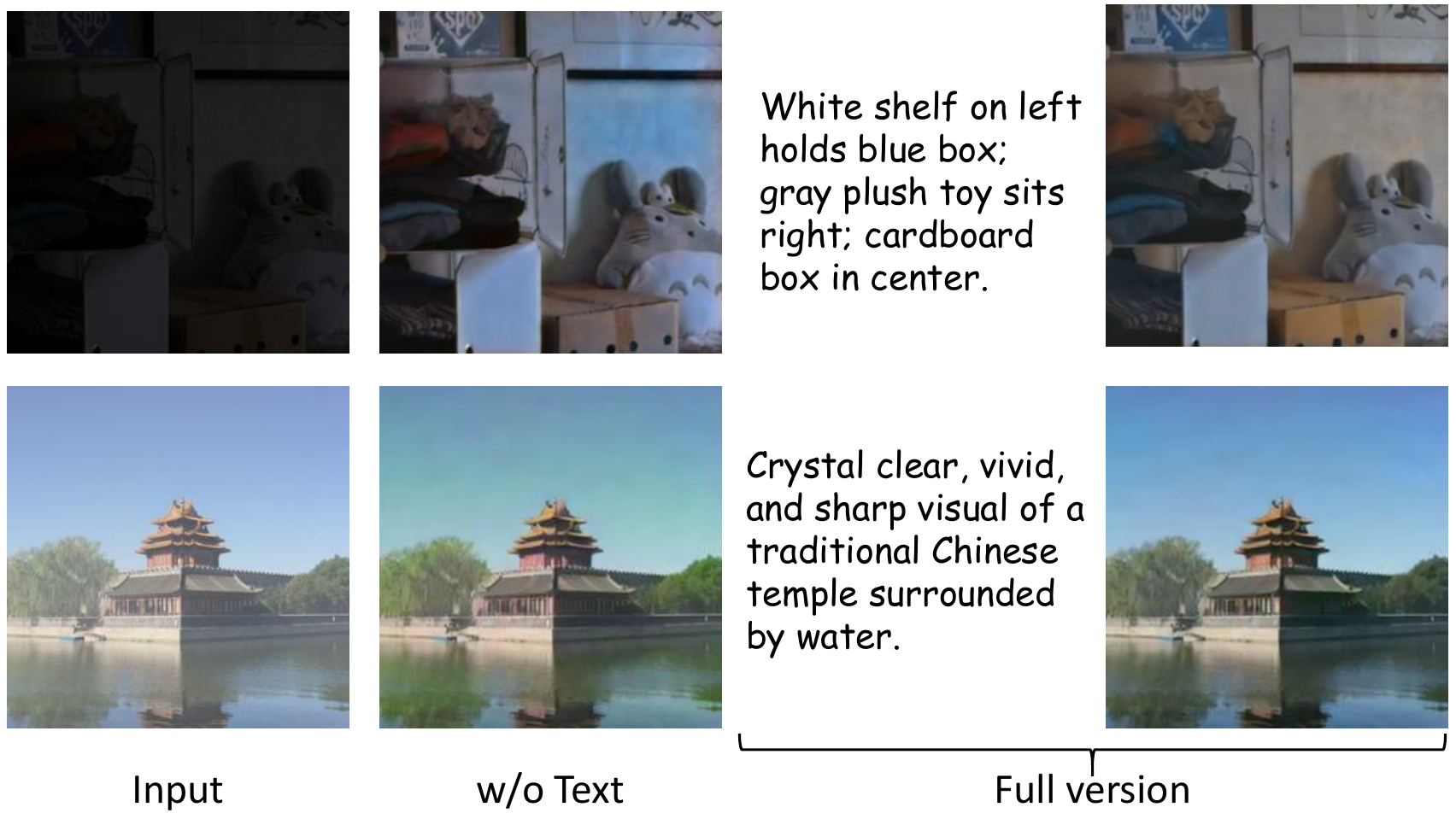}
  \vspace{-2mm}
  \caption{Visualization results of ablation experiments based on guiding text, including enhancement and dehazing tasks.}
  \label{fig:text}
  \vspace{-1.5em}
\end{figure}

\noindent \textbf{Colorizing and correlated mixed degradation.} In the image colorization experiments, we primarily compare our method with GDP. The visualization results are presented in Fig.~\ref{fig:color}. We observe that, although GDP's generative results achieve ideal pixel-level realism, they lack sufficient saturation, causing the images to largely maintain grayscale characteristics. Conversely, our \algoname method facilitates the generation of vibrant colors with enhanced contrast and saturation. Moreover, due to the shortcomings of explicit degradation modeling in accurately capturing the combined degradation of fading and noise, GDP performs poorly in denoising and colorizing gray images. In contrast, our \algoname method effectively mitigates these issues by leveraging the implicit learning of the \moduname.

\subsection{Ablation Study}
To evaluate the mechanism of recurrent refinement in \algoname, we conducted ablation experiments, with the key results presented in Tab. \ref{tab:iter}. The results reveal that the optimal number of recurrences is task-dependent and influenced by the degree of coupling between degradation and semantic features. Specifically, tasks exhibiting stronger feature coupling need more recurrences to adequately preserve semantic information during the restoration process. These results demonstrate that recurrence tuning can effectively balance between degradation reduction and the preservation of semantic content. The supplementary materials also mention that during the recurrence process, the reverse diffusion can better initialize, avoiding the generation of artifacts.

We also performed a series of experiments to evaluate the impact of text guidance in \algoname. As demonstrated in Tab.~\ref{tab:text} and Fig.~\ref{fig:text}, the results reveal that while \algoname can perform restoration tasks without relevant text embeddings, the incorporation of strategically curated textual prompt significantly enhances the diffusion model's ability to comprehend and synthesize desired content. This empirically validates the importance of integrating suitable textual guidance to improve the model performance and output fidelity.

\section{Conclusion}
In this paper, we propose a recurrent posterior sampling method utilizing latent diffusion to tackle the challenge of zero-shot unified image restoration. Our approach leverages the prior knowledge of a pre-trained model to achieve high-quality restoration. Specifically, it learns the degradation patterns from a single image without requiring fine-tuning, thereby exhibiting excellent generalization capabilities. Furthermore, we develop a conditional reverse diffusion based on implicit features and innovatively propose a recurrent refinement strategy to improve the results. Extensive experiments across three categories of methods and five tasks demonstrate the superiority of our \algoname.

\section{Acknowledgements}
This research was funded through National Key Research and Development Program of China (Project No. 2022YFB36066), in part by the Shenzhen Science and Technology Project under Grant (KJZD20240903103210014, JCYJ20220818101001004)

%% file: tab/lolv1.tex
% \usepackage{multirow}
\begin{table*}[htbp]
\centering
\begin{adjustbox}{max width=\textwidth}
\begin{tabular}{l|ccc|ccccc|ccccc}
\noalign{\hrule height 1pt}
\multirow{2}{*}{Methods}       & \multicolumn{3}{c|}{Definition{\rule{0pt}{1.5em}}}                                        & \multicolumn{5}{c|}{LOLv1}                                                                                                           & \multicolumn{5}{c}{LOLv2}                                                                                                           \\ \cline{2-14} 
                               & \multicolumn{1}{l}{B{\rule{0pt}{1.5em}}} & \multicolumn{1}{l}{D} & \multicolumn{1}{l|}{U} & \multicolumn{1}{l}{PSNR$\uparrow$} & \multicolumn{1}{l}{SSIM$\uparrow$} & \multicolumn{1}{l}{LPIPS$\downarrow$} & \multicolumn{1}{l}{PI$\downarrow$} & \multicolumn{1}{l|}{NIQE$\downarrow$} & \multicolumn{1}{l}{PSNR$\uparrow$} & \multicolumn{1}{l}{SSIM$\uparrow$} & \multicolumn{1}{l}{LPIPS$\downarrow$} & \multicolumn{1}{l}{PI$\downarrow$} & \multicolumn{1}{l}{NIQE$\downarrow$} \\ \noalign{\hrule height 1pt}
\textbf{Supervised unified} & \multicolumn{1}{l}{}  & \multicolumn{1}{l}{}  & \multicolumn{1}{l|}{}  & \multicolumn{1}{l}{}     & \multicolumn{1}{l}{}     & \multicolumn{1}{l}{}      & \multicolumn{1}{l}{}   & \multicolumn{1}{l|}{}     & \multicolumn{1}{l}{}     & \multicolumn{1}{l}{}     & \multicolumn{1}{l}{}      & \multicolumn{1}{l}{}   & \multicolumn{1}{l}{}     \\
Airnet~\cite{airnet}                         & \ding{51}                     & \ding{55}                     & \ding{55}                      & --                 &  --                      & --                       & --                     & --                        & --                       & --                       & --                        & --                     & --                       \\
PromptIR~\cite{promptir}                       & \ding{51}                     & \ding{55}                     & \ding{55}                      & --                       & --                       & --                        & --                     & --                        & --                       & --                       & --                        & --                     & --                       \\
DiffUIR~\cite{diffuir}                        & \ding{51}                     & \ding{55}                     & \ding{55}                      & 21.36                    & 0.907                    & 0.125                     & 4.68                   & 5.95                      & 26.14                    & 0.898                    & 0.114                     & 5.26                   & 7.34                     \\ \hline
\textbf{Task-specific}         &                       &                       &                        & \multicolumn{1}{l}{}     & \multicolumn{1}{l}{}     & \multicolumn{1}{l}{}      & \multicolumn{1}{l}{}   & \multicolumn{1}{l|}{}     & \multicolumn{1}{l}{}     & \multicolumn{1}{l}{}     & \multicolumn{1}{l}{}      & \multicolumn{1}{l}{}   & \multicolumn{1}{l}{}     \\
ZDCE++~\cite{zdce}                        & \ding{55}                     & \ding{55}                     & \ding{51}                      & 14.38                    & 0.523                    & 0.240                     & 4.04                   & 4.95                      & 16.76                    & 0.428                    & 0.284                     & 4.17                   & 5.99                     \\
SCI~\cite{sci}                           & \ding{55}                     & \ding{55}                     & \ding{51}                      & 14.86                    & 0.704                    & 0.219                     & 4.42                   & 5.94                      & 17.17                    & 0.639                    & 0.264                     & 6.27                   & 10.47                    \\
CLIP-LIT~\cite{cliplit}                      & \ding{55}                     & \ding{55}                     & \ding{51}                      & 12.63                    & 0.678                    & 0.240                     & 4.28                   & 5.96                      & 15.41                    & 0.650                    & 0.315                     & 6.57                   & 11.31                    \\
ZERO-IG~\cite{zeroig}                        & \ding{55}                     & \ding{51}                     & \ding{51}                      & 17.22                    & 0.794                    & 0.184                     & 4.92                   & 6.22                      & 18.63                    & 0.751                    & 0.231                     & 5.64                   & 8.59                     \\ \hline
\textbf{Posterior sampling}    & \multicolumn{1}{l}{}  & \multicolumn{1}{l}{}  & \multicolumn{1}{l|}{}  & \multicolumn{1}{l}{}     & \multicolumn{1}{l}{}     & \multicolumn{1}{l}{}      & \multicolumn{1}{l}{}   & \multicolumn{1}{l|}{}     & \multicolumn{1}{l}{}     & \multicolumn{1}{l}{}     & \multicolumn{1}{l}{}      & \multicolumn{1}{l}{}   & \multicolumn{1}{l}{}     \\
GDP~\cite{gdp}                            & \ding{55}                     & \ding{51}                     & \ding{51}                      & 16.52                    & 0.690                    & \textbf{0.261}            & \textbf{4.16}          & 5.73                      & 14.48                    & 0.568                    & 0.332                     & 5.31                   & 9.01                     \\
TAO~\cite{TAO}                            & \ding{51}                     & \ding{51}                     & \ding{51}                      & 15.84                    & 0.757                    & 0.363                     & 6.34                   & 8.79                      & 17.63                    & 0.748                    & 0.314                     & 6.24                   & 10.11                    \\
Ours                           & \ding{51}                     & \ding{51}                     & \ding{51}                      & \textbf{17.45}           & \textbf{0.804}           & 0.277                     & 4.79                   & \textbf{5.52}             & \textbf{19.26}           & \textbf{0.751}           & \textbf{0.310}            & \textbf{4.50}          & \textbf{5.57}            \\ \noalign{\hrule height 1pt}
\end{tabular}
\end{adjustbox}
\caption{Comparison results on the LOL datasets, with the highest-performing metrics highlighted in \textbf{bold}. The notation ``--'' indicates that the method is not applicable to this task due to the closed-set nature of supervised methods. To differentiate the generalization capabilities of various methods, we use three distinct characteristics: B, D, U, which represent task-blind, dataset-free, and unsupervised, respectively.}
\label{tab:lol}
\vspace{-0.5em}
\end{table*}

%% file: tab/hsts.tex
% \usepackage{multirow}
\begin{table}[htb]\small
\centering
\begin{adjustbox}{max width=\linewidth}
\begin{tabular}{l|ccc|ccc}
\noalign{\hrule height 1pt}
\multicolumn{1}{c|}{\multirow{2}{*}{\rule{0pt}{1.5em} Methods}} 
& \multicolumn{3}{c|}{\rule{0pt}{1.5em}Defination} 
& \multicolumn{3}{c}{RESIDE} \\
\cline{2-7}
\multicolumn{1}{c|}{}                        & B{\rule{0pt}{1.5em}}                    & D                    & U                     & PSNR$\uparrow$                 & SSIM$\uparrow$                 & LPIPS$\downarrow$                \\ \noalign{\hrule height 1pt}
\textbf{Supervised Unified}                  & \multicolumn{1}{l}{} & \multicolumn{1}{l}{} & \multicolumn{1}{l|}{} & \multicolumn{1}{l}{} & \multicolumn{1}{l}{} & \multicolumn{1}{l}{0} \\
Airnet~\cite{airnet}                                       & \ding{51}                    & \ding{55}                    & \ding{55}                     & 24.37                & 0.899                & 0.059                \\
PromptIR~\cite{promptir}                                     & \ding{51}                    & \ding{55}                    & \ding{55}                     & 25.67                & 0.907                & 0.048                \\
DiffUIR~\cite{diffuir}                                      & \ding{51}                    & \ding{55}                    & \ding{55}                     & 26.88                & 0.914                & 0.045                \\ \hline
\textbf{Task-specific}                       &                      &                      &                       &                      &                      &                      \\
AOD-Net~\cite{aod-net}                                   & \ding{55}                    & \ding{55}                    & \ding{55}                     & 19.15                & 0.860                & 0.129                \\
ZID~\cite{zid}                                          & \ding{55}                    & \ding{51}                    & \ding{51}                     & 19.31                & 0.796                & 0.191                \\
DDIP~\cite{ddip}                                         & \ding{55}                    & \ding{51}                    & \ding{51}                      & 20.20                & 0.846                & 0.150                \\
YOLY~\cite{yoly}                                         & \ding{55}                    & \ding{51}                    & \ding{51}                      & 20.49                & 0.794                & 0.108                \\ \hline
\textbf{Posterior sampling}                  & \multicolumn{1}{l}{} & \multicolumn{1}{l}{} & \multicolumn{1}{l|}{} & \multicolumn{1}{l}{} & \multicolumn{1}{l}{} & \multicolumn{1}{l}{} \\
GDP~\cite{gdp}                                          & \ding{55}                    & \ding{51}                    & \ding{51}                      & 13.15                & 0.757                & 0.144                \\
TAO~\cite{TAO}                                          & \ding{51}                    & \ding{51}                    & \ding{51}                      & 18.38                & \textbf{0.823}       & \textbf{0.147}       \\
Ours                                         & \ding{51}                    & \ding{51}                    & \ding{51}                     & \textbf{21.45}       & 0.813                & 0.177                \\ \noalign{\hrule height 1pt}

\end{tabular}
\end{adjustbox}
\caption{Comparison results on HSTS datasets in RESIDE,  with the highest-performing metrics highlighted in \textbf{bold}.}
\label{tab:hsts}
\vspace{-2em}
\end{table}

%% file: tab/denoising.tex
% \usepackage{multirow}
\begin{table}[htb]\small
\centering
\begin{adjustbox}{max width=\linewidth}
\begin{tabular}{l|ccc|ccc}
\noalign{\hrule height 1pt}
\multicolumn{1}{c|}{\multirow{2}{*}{\rule{0pt}{1.5em} Methods}}  & \multicolumn{3}{c|}{Defination}                                     & \multicolumn{3}{c}{Kodak24}                                                     \\ \cline{2-7} 
\multicolumn{1}{c|}{}                        & B{\rule{0pt}{1.5em}}                    & D                    & U                     & \multicolumn{1}{c}{PSNR$\uparrow$} & \multicolumn{1}{c}{SSIM$\uparrow$} & \multicolumn{1}{c}{LPIPS$\downarrow$} \\ \noalign{\hrule height 1pt}
\textbf{Supervised Unified}                  & \multicolumn{1}{l}{} & \multicolumn{1}{l}{} & \multicolumn{1}{l|}{} &                          &                          &                           \\
Airnet~\cite{airnet}                                       & \ding{51}                    & \ding{55}                    & \ding{55}                     & 29.94                    & 0.834                    & 0.114                     \\
PromptIR~\cite{promptir}                                      & \ding{51}                    & \ding{55}                    & \ding{55}                    & 30.88                    & 0.873                    & 0.113                     \\
DiffUIR~\cite{diffuir}                                       & \ding{51}                    & \ding{55}                    & \ding{55}                    & 22.86                    & 0.789                    & 0.219                     \\ \hline
\textbf{Task-specific}                       &                      &                      &                       & \multicolumn{1}{c}{}     & \multicolumn{1}{c}{}     & \multicolumn{1}{c}{}      \\
Blind2unblind~\cite{b2un}                                & \ding{55}                    & \ding{55}                    & \ding{51}                     & 29.35                    & 0.836                    & 0.141                     \\
ZS-N2N~\cite{zs-n2n}                                       & \ding{55}                    & \ding{51}                    & \ding{51}                     & 30.14                    & 0.862                    & 0.132                     \\
NBR2NBR~\cite{neighbor}                                     & \ding{55}                    & \ding{55}                    & \ding{51}                    & 29.09                    & 0.821                    & 0.119                     \\
Prompt-SID~\cite{prompt-sid}                                  & \ding{55}                    & \ding{55}                    & \ding{51}                     & 30.58                    & 0.866                    & 0.097                     \\ \hline
\textbf{Posterior sampling}                  & \multicolumn{1}{l}{} & \multicolumn{1}{l}{} & \multicolumn{1}{l|}{} &                          &                          &                           \\
GDP~\cite{gdp}                                          & \ding{55}                    & \ding{51}                    & \ding{51}                    & \multicolumn{1}{c}{--}   & \multicolumn{1}{c}{--}   & \multicolumn{1}{c}{--}    \\
TAO~\cite{TAO}                                          & \ding{51}                    & \ding{51}                    & \ding{51}                     & 27.72                    & 0.815                    & 0.179                     \\
Ours                                         & \ding{51}                    & \ding{51}                    & \ding{51}                    & \textbf{28.64}           & \textbf{0.841}           & \textbf{0.175}            \\ \noalign{\hrule height 1pt}
\end{tabular}
\end{adjustbox}
\caption{The comparison results on the Kodak24 dataset is presented, with the highest-performing metrics highlighted in \textbf{bold}. The symbol ``--'' indicates that the method is not applicable to this task due to the lack of relevant priors.}
\label{tab:kodak}
\vspace{-2em}
\end{table}

%% file: tab/iter.tex
% \usepackage{multirow}
\begin{table*}[htbp]\small
\centering
\begin{tabular}{c|ccccc|ccc|ccc}
\noalign{\hrule height 1pt}
\multirow{2}{*}{Recurrence} & \multicolumn{5}{c|}{LOLv1{\rule{0pt}{1.5em}}}                                                                                                           & \multicolumn{3}{c|}{RESIDE}                                                      & \multicolumn{3}{c}{Kodak24}                                                     \\ \cline{2-12} 
                      & \multicolumn{1}{l}{PSNR$\uparrow${\rule{0pt}{1.5em}}} & \multicolumn{1}{l}{SSIM$\uparrow$} & \multicolumn{1}{l}{LPIPS$\downarrow$} & \multicolumn{1}{l}{PI$\downarrow$} & \multicolumn{1}{l|}{NIQE$\downarrow$} & \multicolumn{1}{l}{PSNR$\uparrow$} & \multicolumn{1}{l}{SSIM$\uparrow$} & \multicolumn{1}{l|}{LPIPS$\downarrow$} & \multicolumn{1}{l}{PSNR$\uparrow$} & \multicolumn{1}{l}{SSIM$\uparrow$} & \multicolumn{1}{l}{LPIPS$\downarrow$} \\ \noalign{\hrule height 1pt}
0                     & 16.78{\rule{0pt}{1.0em}}                    & 0.789                    & 0.306                     & 5.33                   & 5.91                      & 19.35                    & 0.779                    & \textbf{0.174}             & 27.75                    & 0.830                    & 0.192                     \\
1                     & 17.21                    & 0.792                    & 0.295                     & 5.18                   & 5.99                      & 20.38                    & 0.807                    & 0.181                      & \textbf{28.60}           & \textbf{0.841}           & \textbf{0.171}            \\
2                     & \textbf{17.73}           & \textbf{0.807}           & \textbf{0.287}            & 4.98                   & 5.71                      & 20.83                    & 0.806                    & 0.185                      & 28.26                    & 0.842                    & 0.185                     \\
3                     & 17.10                    & 0.806                    & 0.288                     & \textbf{4.74}          & \textbf{5.52}             & \textbf{21.60}           & \textbf{0.810}           & 0.179                      & 28.49                    & 0.842                    & 0.186                     \\ \noalign{\hrule height 1pt}
\end{tabular}
\vspace{-2mm}
\caption{Ablation experiments on the number of recurrences for \algoname conducted on the LOLv1 dataset, the HSTS subset of the RESIDE dataset, and the Kodak24 dataset, with the best results highlighted in \textbf{bold}.}
\label{tab:iter}
\vspace{-2mm}
\end{table*}

%% file: tab/text.tex
% \usepackage{multirow}
\begin{table*}[htbp]\small
\centering
\begin{tabular}{l|ccccc|ccc|ccc}
\noalign{\hrule height 1pt}
\multirow{2}{*}{Methods} & \multicolumn{5}{c|}{LOLv1{\rule{0pt}{1.5em}}}                                                                                                           & \multicolumn{3}{c|}{RESIDE}
& \multicolumn{3}{c}{Kodak24}
\\ \cline{2-12} 
                         & \multicolumn{1}{l}{PSNR$\uparrow${\rule{0pt}{1.5em}}} & \multicolumn{1}{l}{SSIM$\uparrow$} & \multicolumn{1}{l}{LPIPS$\downarrow$} & \multicolumn{1}{l}{PI$\downarrow$} & \multicolumn{1}{l|}{NIQE$\downarrow$} & \multicolumn{1}{l}{PSNR$\uparrow$} & \multicolumn{1}{l}{SSIM$\uparrow$} & \multicolumn{1}{l|}{LPIPS$\downarrow$}& \multicolumn{1}{l}{PSNR$\uparrow$} & \multicolumn{1}{l}{SSIM$\uparrow$} & \multicolumn{1}{l}{LPIPS$\downarrow$} \\ \noalign{\hrule height 1pt}
w/o Text{\rule{0pt}{1em}}                 & 16.03                    & 0.774                    & 0.312                     & 5.33                   & 5.82                      & 19.63                    & 0.791                    & \textbf{0.169}&28.13&0.838&0.179                     \\
Full Version             & \makecell{\textbf{17.73}\\(+1.70)}             & \makecell{\textbf{0.807}\\(+0.03)}             & \makecell{\textbf{0.288}\\(-0.02)}              & \makecell{\textbf{4.98}\\(-0.35)}            & \makecell{\textbf{5.71}\\(-0.11)}               & \makecell{\textbf{21.60}\\(+1.97)}             & \makecell{\textbf{0.810}\\(+0.02)}             & \makecell{0.179\\(-0.01)} & \makecell{\textbf{28.60}\\(+0.47)}& \makecell{\textbf{0.841}\\(+0.00)}& \makecell{\textbf{0.171}\\(-0.01)}             \\ \noalign{\hrule height 1pt}
\end{tabular}
\vspace{-2mm}
\caption{Experiments on \algoname incorporating classifier guidance based on textual descriptions, with the best results highlighted in \textbf{bold}. Additionally, improvements brought by using text embeddings are noted.}
\label{tab:text}
\vspace{-1em}
\end{table*}